\documentclass[runningheads]{llncs}
\usepackage[T1]{fontenc}
\usepackage{graphicx}
\usepackage{amsmath}
\begin{document}
\title{Measuring What Matters---or What’s Convenient?: Robustness of LLM-Based Scoring Systems to Construct-Irrelevant Factors}
\titlerunning{Robustness of LLM-Based Scoring}
%
\author{Cole Walsh\orcidID{0000-0002-6284-8926} \and
Rodica Ivan\orcidID{0000-0001-6031-7200}}
\authorrunning{C. Walsh and R. Ivan}
%
\institute{Acuity Insights Inc., Toronto, ON, Canada\\
\email{\{cwalsh,rivan\}@acuityinsights.com}}
\maketitle              
\begin{abstract}
Automated systems have been widely adopted across the educational testing industry for open-response assessment and essay scoring. These systems commonly achieve performance levels comparable to or superior than trained human raters, but have frequently been demonstrated to be vulnerable to the influence of construct-irrelevant factors (i.e., features of responses that are unrelated to the construct assessed) and adversarial conditions. Given the rising usage of large language models in automated scoring systems, there is a renewed focus on ``hallucinations'' and the robustness of these LLM-based automated scoring approaches to construct-irrelevant factors. This study investigates the effects of construct-irrelevant factors on a dual-architecture LLM-based scoring system designed to score short essay-like open-response items in a situational judgment test. It was found that the scoring system was generally robust to padding responses with meaningless text, spelling errors, and writing sophistication. Duplicating large passages of text resulted in lower scores predicted by the system, on average, contradicting results from previous studies of non-LLM-based scoring systems, while off-topic responses were heavily penalized by the scoring system. These results provide encouraging support for the robustness of future LLM-based scoring systems when designed with construct relevance in mind.

\end{abstract}

\section{Introduction}

Automatic evaluation of open-response text, including short-answer responses and essays, is one of the earliest and most widely explored applications of natural language processing and artificial intelligence (AI) in education. Over time, methods for automatically evaluating written work have evolved from using handcrafted features (e.g., type-token ratios, part-of-speech tagging) and simple models~\cite{klebanov2020automated} to more complex approaches including neural networks and transformer-based models~\cite{klebanov2020automated,ravindran-choi-2025-investigating}. With each new development automated scoring systems have consistently demonstrated strong alignment with human raters~\cite{klebanov2020automated}.

Despite their strong overall performance and successful adoptions, studies have demonstrated the susceptibility of many automated scoring systems to construct-irrelevant factors (i.e., features of responses that are unrelated to the construct assessed) regardless of underlying architecture. Chief among these construct-irrelevant factors is text length; early studies involving simple models demonstrated that repeating sentences or entire paragraphs within an essay could artificially inflate scores predicted by these models~\cite{powers2002stumping}, while more recent studies have demonstrated that this issue persists even in transformer-based scoring systems~\cite{ravindran-choi-2025-investigating}. Other studies have demonstrated outsized impacts of injecting particular words or phrases (e.g., ``from one perspective'', ``on the other hand'', ``in the final analysis'')~\cite{powers2002stumping} and including off-topic text~\cite{powers2002stumping,zhang2016evaluating}. The fragility of existing automated scoring systems under such adversarial conditions undermines trust in these systems to accurately measure the intended construct rather than exploitable proxies.

In recent years, automated scoring research has begun to focus on systems employing large language models (LLMs) either directly~\cite{lee2024applying} or as part of more complex scoring systems~\cite{bruno-becker-2025-explainable,walsh-etal-2025-using}. While these systems have not yet met state-of-the-art performance metrics achieved by transformer-based systems in many cases~\cite{bruno-becker-2025-explainable,huang-wilson-2025-evaluating}, what they sacrifice in accuracy they make up for in other ways: creating human-interpretable scores and feedback, reducing the amount of required training data, and easing adoption by less technical assessment developers. With this rise in LLM-based scoring solutions, there is a renewed focus on the robustness of these systems under adversarial conditions, especially with widespread public awareness of the limitations in the underlying technology (e.g., ``hallucinations''~\cite{farquhar2024detecting}), which was not the case with prior modes of automated scoring.

This study contributes towards our understanding of the effects of construct-irrelevant factors on LLM-based scoring systems by investigating how the following factors influence scores produced by one such system:
\begin{enumerate}
    \item Meaningless Text,
    \item Writing Sophistication, and
    \item Off-Topic Texts
\end{enumerate}
Writing sophistication is often considered a construct-relevant factor by assessments of writing or language proficiency~\cite{bruno-becker-2025-explainable,naismith2025duolingo}. As will be discussed in greater detail in Sec.~\ref{sec:assessment}, we constructed our LLM-based scoring system for a situational judgment test (SJT) designed to measure personal and professional skills (e.g., communication, teamwork, problem-solving, and critical thinking)~\cite{mcdaniel2007situational} where elements of writing sophistication (e.g., spelling, grammar, structure, and organization) were considered construct-irrelevant. This study, then, serves a secondary purpose of investigating the feasibility of constructing automated scoring systems that measure distinctly different constructs from scoring systems employed for measuring writing or language proficiency.

\section{Methods}

\subsection{Source Data}

\subsubsection{Assessment}\label{sec:assessment}

We collected data using a 30 item open-response SJT designed to assess and provide feedback on students' personal and professional skills. This instrument, intended primarily for low-stakes formative (as opposed to summative) usage within a program of study, assessed students' skills along four dimensions~\cite{iqbal_formative}:
\begin{enumerate}
    \item \textbf{Intrapersonal skills:} Understanding and regulating oneself by recognizing emotions, biases, and behaviors, staying motivated, adapting to challenges, and committing to lifelong learning for personal growth.
    \item \textbf{Interpersonal skills:} Building meaningful connections and working effectively with others by communicating clearly, showing empathy, collaborating toward shared goals, and inspiring others towards positive outcomes.
    \item \textbf{Social and Ethical Responsibility}: Recognizing and respecting people’s differences, upholding ethical principles, and contributing to the well-being of society through responsible actions.
    \item \textbf{Critical Thinking and Problem Solving:} Gathering information, evaluating options, and finding effective solutions to problems while efficiently managing resources and risks.
\end{enumerate}

For each item, students were presented with a hypothetical scenario and asked how they would respond given the situation. Scenarios were presented in one of two formats: \textit{text-based} scenarios included a short description of a situation, while \textit{video-based} scenarios were depicted using AI-generated avatars enacting a situation. Below is a summary of one text-based scenario students were shown and an accompanying item (question):
\begin{quote}
    \textbf{Scenario:} Kendra and Alex, two close friends co-leading a major group project, are clashing over Kendra pushing a strict schedule to meet deadlines. Alex wants more group input on task division and worries that Kendra’s approach could undermine team trust and strain their friendship.\\
    \textbf{Item:} How could the co-leads, Alex and Kendra, have collaborated more effectively with the group to create a clear and realistic project plan?
\end{quote}
This item was meant to assess interpersonal skills and, more specifically, collaboration. To receive a high score on this item, students were expected to demonstrate behaviors consistent with collaboration in their responses such as \textit{suggesting approaches to managing differences and resolving conflict}.

\subsubsection{Automated Scoring System}

We developed an automated scoring system for this assessment that used a dual-architecture LLM-as-a-Judge feature extraction component together with clear-box regression algorithms similar to those described in Refs.~\cite{bruno-becker-2025-explainable,walsh-etal-2025-using}. Aligning with the constructs assessed by the instrument, we designed the scoring system to reward higher-level features related to personal and professional skills (such as the collaboration-related behavior noted above) rather than language proficiency or content-mastery. We constructed this system using $26, 571$ assessment responses (spread across all 30 items) from 910 students representing six schools. These schools comprised a mixture of business, engineering, health sciences, and medicine programs. All responses were first evaluated by trained human raters using a 1--5 Likert rating scale; these ratings were used to train and evaluate the automated scoring system, which achieved performance comparable to the human raters.

\subsection{Data Selection}

For this study, we selected a subset of 545 responses (spread across all 30 items) from 318 students. We chose to use a subset of responses rather than the full dataset described above to minimize LLM API costs given that we would be executing a number of experiments that involved re-scoring responses. Through simulations with varying sample sizes, we identified that a set of at least 500 responses would allow us to reliably calculate paired Cohen's $d$ effect sizes with our desired precision (width of 95\% confidence interval $< 0.2$); we developed our sampling strategy with this heuristic in mind. To ensure all assessment items, as well as responses of varying quality, were represented in the sampled dataset, we selected responses from our larger dataset by stratifying by item and predicted score: we binned predicted scores using 10 equal width bins between the available scoring range of 1--5, then selected up to two responses at random from each scoring bin for each item. This sampling strategy ensured that we selected a diversity of low and high quality responses from all 30 assessment items. Our automated scoring system did not produce scores in certain scoring bins for certain items (e.g., the model may not have predicted any scores in the (1.4, 1.8] bin for an item), hence our sampled dataset included less than the total of 600 responses that would have been expected had all scoring bins been attainable for all 30 items. \textit{Table~\ref{tab:bin_breakdown}} includes a breakdown of the number of responses selected for each scoring bin. The mean predicted score for sampled responses was $3.11$ with a standard deviation of $1.17$.

\begin{table}
\centering
\caption{Number of responses selected from each scoring bin. Our automated scoring system did not produce scores in certain scoring bins for certain items, hence not all scoring bins included the maximum 60 responses.}\label{tab:bin_breakdown}
\begin{tabular}{| c | c |}
\hline
\textbf{Scoring Bin} & \textbf{N} \\
\hline
[1, 1.4] & 60 \\
(1.4, 1.8] & 42 \\
(1.8, 2.2] & 48 \\
(2.2, 2.6] & 50 \\
(2.6, 3.0] & 55 \\
(3.0, 3.4] & 56 \\
(3.4, 3.8] & 55 \\
(3.8, 4.2] & 60 \\
(4.2, 4.6] & 60 \\
(4.6, 5.0] & 59 \\
\hline
\end{tabular}
\end{table}

Students consented to the use of their assessment and survey data for research purposes by accepting the \textit{Terms and Conditions} before starting the assessment; students were informed that they may withdraw their consent at any time. Only students who consented to their data being used for research purposes were included in this study. All students who completed the assessment could also complete an optional demographic information survey at the end of the test. \textit{Table~\ref{tab:demographic_breakdown}} provides a breakdown of the self-identified demographic characteristics for the students whose responses we selected for this study. 

\begin{table}
\centering
\caption{Number of students included in this study from various demographic subgroups.  To protect individual privacy, any subgroups with fewer than 5 individuals were included in `Other' subcategories.}\label{tab:demographic_breakdown}
\begin{tabular}{|l c c|}
\hline
\textbf{Demographic Characteristic} & \textbf{N} & \textbf{Fraction}\\
\hline
\textit{English Proficiency} & & \\
\hspace{1mm} Other & 5 & 1.6\% \\
\hspace{1mm} Good & 14 & 4.4\% \\
\hspace{1mm} Advanced & 44 & 13.8\% \\
\hspace{1mm} Native/Funationally Native & 255 & 80.2\% \\
\textit{Gender} & & \\
\hspace{1mm} Other & 1 & 0.3\% \\
\hspace{1mm} Woman & 149 & 46.9\% \\
\hspace{1mm} Man & 168 & 52.8\% \\
\textit{Race} & & \\
\hspace{1mm} Other & 4 & 1.3\% \\
\hspace{1mm} Southeast Asian & 12 & 3.8\% \\
\hspace{1mm} Middle Eastern or Northern African & 13 & 4.1\% \\
\hspace{1mm} East Asian & 20 & 6.3\% \\
\hspace{1mm} Black, African, Caribbean, or African American & 29 & 9.1\% \\
\hspace{1mm} South Asian & 31 & 9.7\% \\
\hspace{1mm} Hispanic, Latinx, or Spanish origin & 46 & 14.5\% \\
\hspace{1mm} White or European & 163 & 51.3\% \\
\hline
\end{tabular}
\end{table}

\subsection{Experiments}

We conducted three experiments investigating the influence of construct-irrelevant factors on scores predicted by our automated scoring system:
\begin{enumerate}
    \item Adding meaningless text that does not add to the content of the response
    \item Varying writing sophistication in terms of spelling errors, vocabulary, and sentence structure
    \item Generating off-topic responses that are not aligned with the item
\end{enumerate}
We describe each experiment in greater detail below. In each case, we produced new datasets of 545 responses with each generated response as an altered version of one of the original responses from our base dataset. We re-scored the altered responses using our automated scoring system, then computed statistics over the newly computed scores including paired Cohen's $d$ effect size estimates using the original predicted scores as a baseline for the paired comparisons.

\subsubsection{Experiment 1: Meaningless Text}

We tested the impact of appending four types of meaningless text to the original response:
\begin{enumerate}
    \item[A.] A copy of the original response
    \item[B.] A sentence stating what competency was assessed by the item (e.g., ``This question is designed to assess collaboration.'')
    \item[C.] A sentence rephrasing what happened in the scenario. See example in Sec.~\ref{sec:assessment}.
    \item[D.] A formulaic sentence: ``I would approach the situation in a respectful, non-confrontational, and non-judgmental manner.''
\end{enumerate}
While sub-experiments B and C introduced no meaningful additional content to the original responses, sub-experiment A was meant to explore the particularly adversarial condition where the same text is repeated, inflating response length without any additional content. Ravindran \& Choi previously identified that this exact alteration led to an average change in predicted scores of $0.93$ (on a 1--6 scale) using a transformer-based scoring system~\cite{ravindran-choi-2025-investigating}, so we wanted to explore whether this result translated to LLM-based scoring systems. Sub-experiment D does introduce meaningful text, but is emblematic of the kind of memorized, formulaic phrases often observed in this and other open-response SJTs~\cite{iqbal2025evaluating}.

\subsubsection{Experiment 2: Writing Sophistication}\label{sec:writing_soph}

We tested two types of alterations to the original responses affecting the overall written quality and sophistication of the responses:
\begin{enumerate}
    \item[A.] Introducing spelling errors
    \item[B.] Adjusting the reading level needed to understand the response
\end{enumerate}
For sub-experiment A, we introduced random character-level edits to responses with fixed probabilities ranging from 5--50\% in steps of five percentage points. We used a fixed distribution of edit types with 40\% of edits being substitutions, 30\% being deletions, and 30\% being insertions. For substitutions and insertions, inserted characters were selected from all upper and lowercase english characters with equal weighting. This procedure produced a ``worst-case'' result as there was no predictable pattern in generated errors unlike those typically seen in human writing~\cite{horbach2017influence}; the results of this experiment should therefore be interpreted as a lower bound on the robustness of the scoring system to spelling errors. Previous studies of LLM accuracy for more general tasks found disparities in the robustness of LLMs to these types of errors along the lines of model capability, with more advanced models generally showing superior robustness to spelling errors in a prompt~\cite{gan2024reasoning}. The scoring system we investigated here used GPT-5.X models, so we hypothesized that it would be similarly robust to spelling errors.

For sub-experiment B, we used an LLM (GPT-5 mini) to rephrase each response at a higher or lower grade reading level while maintaining the original meaning of the response. We measured the reading level of the original and altered responses using the Flesch–Kincaid Reading Grade Level (RGL)~\cite{thomas1975test}. This index considers texts with more syllables per word and more words per sentence as more difficult to read and requiring more education to understand. While an imperfect measure of writing sophistication, this index provides a useful heuristic for measuring low-level differences in text structures, allowing us to investigate whether and to what extent more complex words and sentences influence predicted scores.

Features of writing quality, such as these, are construct-relevant for traditional writing and language proficiency assessments (see, for example, Refs.~\cite{bruno-becker-2025-explainable,naismith2025duolingo}), but not SJTs. This experiment is particularly important, then, for gauging whether our automated scoring system is effectively evaluating the intended constructs and not constructs related to writing proficiency.

\subsubsection{Experiment 3: Off-Topic Responses}

In this experiment we investigated how off-topic responses (i.e., responses that did not address the scenario and question presented to the student) were interpreted by our automated scoring system. To do this we created random permutations of item-response combinations from our original dataset such that each response was still represented exactly once and was not matched with its original item. We conducted two versions of this experiment:
\begin{enumerate}
    \item[A.] Responses were matched with an item assessing a \textit{different} competency as the original item
    \item[B.] Responses were matched with an item assessing the \textit{same} competency as the original item
\end{enumerate}
Whereas for items assessing the same competency (e.g., collaboration), our automated scoring system was designed to evaluate the same underlying features (though potentially weight them differently in scoring), there was no overlap between features evaluated in items assessing different competencies. We therefore hypothesized that off-topic responses that still exhibited relevant features expected of high quality responses for the competency would receive higher scores than off-topic responses that exhibited features unrelated to the competency.

\section{Results and Discussion}

\subsection{Experiment 1: Meaningless Text}

\textit{Table~\ref{tab:non-content_text}} shows the mean word count of the responses evaluated in each sub-experiment as well as the mean and standard deviation (SD) of scores assigned by our automated scoring system. We also include the the baseline (unaltered) dataset for comparison and the paired Cohen's $d$ effect size estimate between scores assigned to the baseline responses and responses in each sub-experiment. Contrary to the findings by Ravindran \& Choi~\cite{ravindran-choi-2025-investigating}, we find that simply duplicating a response generally \textit{decreases} the score assigned by our automated scoring system by a small amount (Cohen's $d = -0.24$). On the other hand, adding other kinds of meaningless text like a statement of the competency tested by the item (sub-experiment B) or a rephrasing of the scenario (sub-experiment C) has negligible impact on scores ($|d| \leq 0.01$). We did find, however, that adding certain formulaic phrases (sub-experiment D) could positively bias our scoring system to assign higher scores, though (at least in the case examined here) those effects are very small ($d = 0.16$). 

\begin{table}
\centering
\caption{Summary of responses evaluated in our baseline dataset and each sub-experiment in Experiment 1 including mean word count of the responses evaluated, the mean and standard deviation (SD) of scores assigned by our automated scoring system, and paired Cohen's $d$ effect size estimates between scores assigned in each sub-experiment with the baseline dataset.}\label{tab:non-content_text}
\begin{tabular}{| c | c | c | c |}
\hline
\textbf{Sub-Experiment} & \textbf{Mean Word Count} & \textbf{Mean Score (SD)} & \textbf{Cohen's $d$ (95\% CI)}\\
\hline
Baseline & 54.0 & 3.11 (1.17) & --- \\
A & 107.4 & 2.82 (1.12) & -0.24 (-0.33, -0.16) \\
B & 61.6 & 3.09 (1.15) & -0.01 (-0.10, 0.07) \\
C & 95.7 & 3.12 (1.14) & 0.01 (-0.08, 0.09) \\
D & 66.0 & 3.28 (1.02) & 0.16 (0.08, 0.24) \\
\hline
\end{tabular}
\end{table}

These results indicate that increasing text verbosity without meaningfully adding to the content of the text is unlikely to positively influence scores assigned by our automated scoring system. By adding a statement about the competency tested by the item or a rephrasing of the scenario presented in the item we were able to increase mean response length by 14\% and 77\%, respectively, with effectively zero change in the mean scores assigned. Adding text that detracted from the clarity of the overall response by duplicating the original response even led to negative overall shifts in scores assigned, again pointing to the importance of precise responses over those that are extraneously verbose. Adding text to responses can positively influence scores assigned by our scoring system, but only when that text contains meaningful and new information, as we found in sub-experiment D. The results of that sub-experiment indicate that adding memorized, formulaic phrases may only have very small effects on assigned scores, though more in-depth investigation of the use of such phrases on their own and in combination would be needed to understand the full scope of the effects.

\subsection{Experiment 2: Writing Sophistication}

\subsubsection{Spelling Errors}

\begin{figure}
\includegraphics[width=\textwidth]{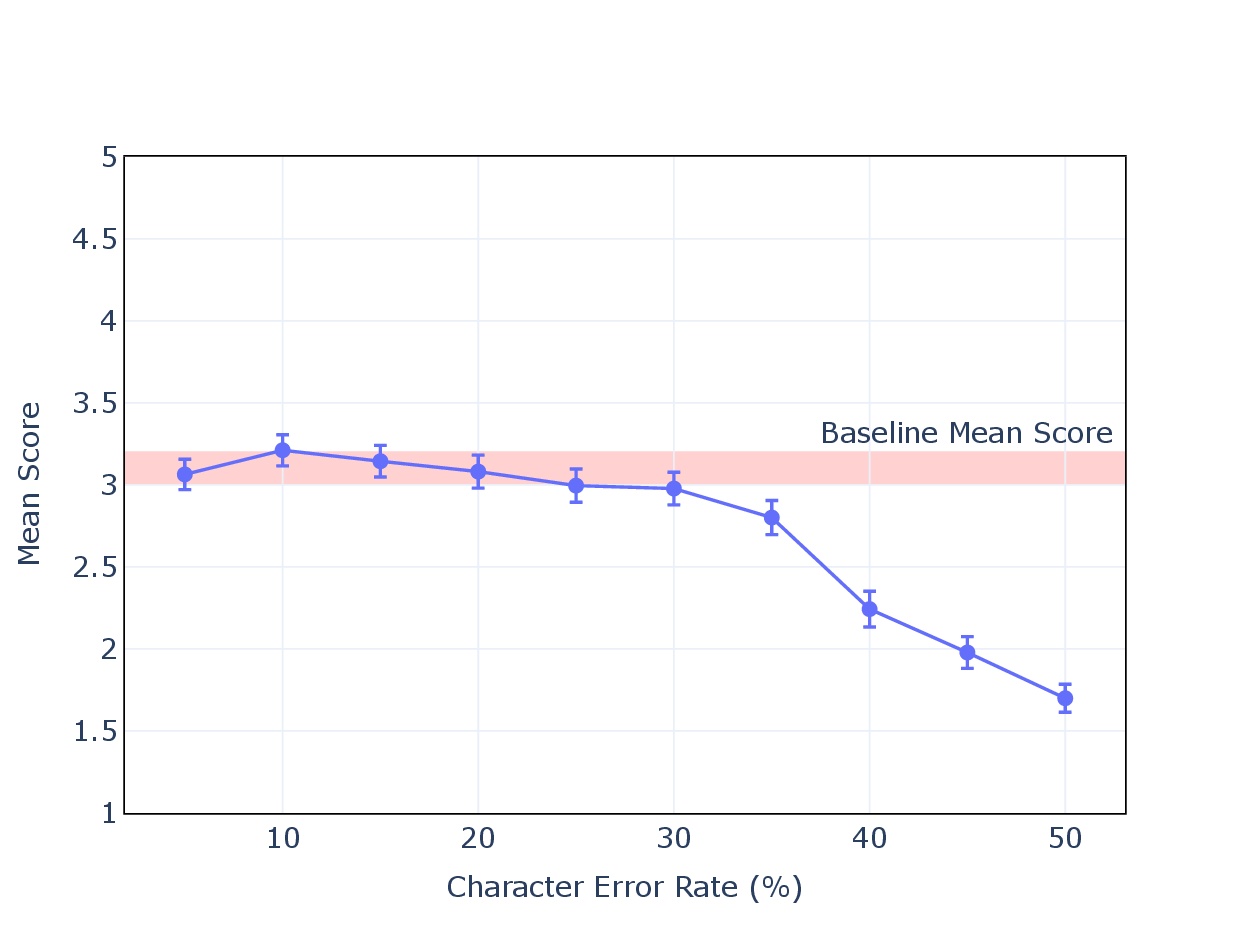}
\caption{Mean score predicted by our automated scoring system for simulations of different character error rates (CER) in responses. The effects of spelling errors on the scoring system only become meaningful at CERs of 35\% and above. Error bars represent the 95\% confidence interval for the mean predicted score. The shaded region represents the 95\% confidence interval on the mean score assigned to responses in the baseline dataset.} \label{fig:scores_cer}
\end{figure}

\textit{Table~\ref{tab:score_cer}} shows the mean score assigned to responses in each simulation with different character error rates (CERs). We also include the baseline (unaltered) dataset for comparison and the paired Cohen’s d effect size estimate between scores assigned to the baseline responses and responses in each simulation. We find very small or negligible effects of spelling errors up to a CER of about 30\% with small observable effects showing up at a CER of 35\%. Large or very large effects appear at CERs of 40\% and beyond as the texts begin to become indistinguishable from random strings. \textit{Fig.~\ref{fig:scores_cer}} illustrates the average score predicted by our automated scoring system as a function of CER as well, where this drop-off in mean predicted scores is more clearly observable.

\begin{table}
\centering
\caption{Summary of simulations introducing spelling errors into responses. The character error rate (CER) denotes the probability of any character in a response being edited (substitution, deletion, insertion). Compared to our baseline dataset with no artificial spelling errors introduced, there are very small or negligible differences in scores assigned by our automated scoring system, on average, up to a CER of about 30\%.}\label{tab:score_cer}
\begin{tabular}{| c | c | c |}
\hline
Character Error Rate & Mean Predicted Score (SD) & Cohen's $d$ (95\% CI)\\
\hline
Baseline & 3.11 (1.17) & --- \\
5\% & 3.06 (1.10) & -0.04 (-0.12, 0.05) \\
10\% & 3.21 (1.13) & 0.09 (0.01, 0.18) \\
15\% & 3.14 (1.15) & 0.03 (-0.05, 0.12) \\
20\% & 3.08 (1.20) & -0.02 (-0.10, 0.06) \\
25\% & 3.00 (1.21) & -0.09 (-0.18, -0.01) \\
30\% & 2.98 (1.18) & -0.11 (-0.19, -0.2) \\
35\% & 2.80 (1.24) & -0.25 (-0.34, -0.17) \\
40\% & 2.24 (1.29) & -0.70 (-0.79, -0.60) \\
45\% & 1.98 (1.15) & -0.97 (-1.07, -0.87) \\
50\% & 1.70 (1.01) & -1.28 (-1.40, -1.17) \\
\hline
\end{tabular}
\end{table}

These results indicate that our scoring system is robust to spelling errors. To illustrate, simulating a 30\% CER in the previous sentence produces the following result: ``These esflts indicape thst ofur ecsrind sytv ails rnobstOo speloiVg errgprs.'' This altered sentence may be barely legible for a human reader, but the LLMs used as part of our scoring system appear to be able to interpret these responses sufficiently in regards to the provided instructions. This is a desired behavior of our assessment and automated scoring system as it allows students to not focus on spelling while completing the assessment.

\subsubsection{Reading Level}

\textit{Table~\ref{tab:score_rl}} provides a summary of six simulation studies adjusting the RGL of the original responses to varying degrees. While the word count remains relatively stable across simulations, the average number of words per sentence increases progressively over the simulations as we alter the responses to include fewer, but longer and more complex, sentences. We also observe that the average number of syllables per word increases steadily over the simulations as we replace simpler words with more complex ones. Combined, these alterations produce responses with higher RGLs, on average.

\begin{table}
\centering
\caption{Summary of simulations adjusting original responses to reflect different Flesch-Kincaid Reading Grade Levels (RGLs), which accounts for the average number of words per sentence and syllables per word in a text. Texts with higher RGLs can be interpreted as requiring more education to understand. Results from the baseline dataset where responses were not altered are italicized. We find very small or negligible differences in scores assigned to sets of responses with mean RGLs spanning up to six points.}\label{tab:score_rl}
\begin{tabular}{| c | c | c | c | c | c | c |}
\hline
\multicolumn{6}{|c|}{\textbf{Mean Value}} & \\
\cline{1-6}
\rule{0pt}{4mm}\textbf{RGL} & \textbf{N(words)} & \textbf{N(sentences)} & $\frac{\text{\textbf{words}}}{\text{\textbf{sentence}}}$ & $\frac{\text{\textbf{syllables}}}{\text{\textbf{word}}}$ & \textbf{Score (SD)} & $\mathbf{d}$ \textbf{(95\% CI)} \rule[-2mm]{0pt}{0pt} \\
\hline
3.2 & 42.3 & 4.2 & 10.4 & 1.2 & 2.89 (1.13) & -0.19 (-0.27, -0.10) \\
4.7 & 47.1 & 3.7 & 12.9 & 1.3 & 3.00 (1.14) & -0.09 (-0.17, 0.00) \\
6.5 & 48.9 & 3.2 & 15.6 & 1.3 & 3.05 (1.14) & -0.05 (-0.14, 0.03) \\
7.6 & 48.2 & 2.9 & 16.8 & 1.4 & 3.10 (1.16) & -0.01 (-0.09, 0.08) \\
8.6 & 47.7 & 2.7 & 17.8 & 1.5 & 3.08 (1.14) & -0.02 (-0.11, 0.06) \\
\textit{10.5} & \textit{54.0} & \textit{2.6} & \textit{21.7} & \textit{1.5} & \textit{3.11 (1.17)} & --- \\
13.2 & 48.9 & 2.4 & 20.9 & 1.8 & 3.22 (1.15) & 0.10 (0.02, 0.19) \\
\hline
\end{tabular}
\end{table}

We find very small or negligible differences in predicted scores for sets of responses with mean RGLs up to roughly six points apart. For instance, responses in our baseline dataset had a mean RGL of $10.5$, while in one simulation we produced altered responses with mean RGL of $4.7$; the paired Cohen's $d$ effect size between scores assigned to these two sets of responses was $0.09$. In two other simulations we produced sets of responses with mean RGLs of $7.6$ and $13.2$; the paired Cohen's $d$ effect size between scores assigned to these two sets of responses was $0.11$. We can illustrate how differences in RGLs manifest in responses with an example; below is a sample response to the item from Sec.~\ref{sec:assessment} with an RGL of $11.7$:
\begin{quote}
    \textit{Alex and Kendra could have combined structure with input by first outlining a rough timeline and then holding a short group meeting to adjust it together. This would keep the project moving while still allowing members to feel involved. Kendra’s focus on deadlines and Alex’s concern about trust are both valid, and acknowledging that openly could have helped balance efficiency with collaboration.}
\end{quote}
In one simulation, we edited this response to an RGL of $2.5$:
\begin{quote}
    \textit{Alex and Kendra could make a plan and ask the group to help. First they can make a simple plan with dates. Then they can have a short team meeting to change the plan together. This keeps the work moving and lets people help. Kendra cares about finish dates. Alex worries about trust. Both are right. Saying this out loud can help them work fast and work together.}
\end{quote}
The edited response is slightly longer (68 words compared to 62 in the original), contains simpler sentences, and does away with more complex words (e.g., ``outlining'', ``allowing'', ``acknowledging'', ``efficiency'', ``collaboration''). Despite these changes, the edited response still maintains the tone and content of the original response; our automated scoring system assigned a score of $4.6$ to both responses.

These results indicate that low-level writing features like sentence structure and vocabulary are largely inconsequential to our automated scoring system. Taken together with the observed robustness of our scoring system to spelling errors noted above, we have strong evidence that writing quality and sophistication are generally ignored by our system. Given that automated scoring systems to date have predominantly been designed to measure exactly these kinds of features, this result affirms that our system is indeed capturing something different, distinguishing it from other scoring systems measuring writing or language proficiency.

\subsection{Experiment 3: Off-Topic Responses}

We find that off-topic responses are appropriately penalized by our scoring system. Responses matched with a different, random item assessing a different competency (e.g., a response to an item targeting \textit{collaboration} was matched with a different random item \textit{not} targeting \textit{collaboration}) received an average score of $1.69$ compared to the same responses when correctly matched with their original items, which received an average score of $3.11$ (Cohen's $d = -1.33$). Responses matched with a different, random item assessing the same competency (e.g., a response to an item targeting \textit{collaboration} was matched with a different random item \textit{also} targeting \textit{collaboration}) received scores of $2.38$, on average (Cohen's $d = -0.62$ compared to scores assigned to responses correctly matched with their original items). Our scoring system also returned the minimum possible score of $1.0$ more often when the response was not aligned with the item. In our baseline dataset, our scoring system assigned 20 scores of $1.0$ (3.7\%); this number rises to 72 (13.2\%) when responses are matched with different items assessing the same competency and 191 (35.0\%) when matched with different items assessing a different competency.

These results provide two main insights. First, an otherwise strong response that is misaligned with the task is considered unfavorably by our scoring system and is likely to result in the lowest attainable score. Second, our scoring system appropriately considers different competencies in responses for different items. If our scoring system rewarded similar features of responses across different items we would have expected to see similar results in both versions of the experiment conducted here. That we did not points to our scoring system working as expected: it rewards certain features of responses for certain items and other features for other items.

\section{Conclusion}

This study investigated the susceptibility of an LLM-based automated scoring system to construct-irrelevant factors. Our results indicated that the scoring system investigated was robust to padding response length with meaningless text either in the form of stating what competency was assessed by an item or rephrasing the scenario prompt provided. We also found that duplicating text generally had deleterious effects on scores predicted by the scoring system, in contrast to results from previous studies of non-LLM-based scoring systems~\cite{powers2002stumping,ravindran-choi-2025-investigating} where duplicated text inflated predicted scores. Given these results, LLM-based scoring solutions may offer an avenue for overcoming limitations of existing systems that over-index on text length.

We also found that the investigated scoring system was relatively robust to writing sophistication. Spelling errors only began to noticeably affect scores predicted by the system at character error rates greater than 30\%, a result aligned with previous studies of robustness of automatic content scoring (not writing or language proficiency)~\cite{horbach2017influence}. Vocabulary and response structure, similarly, had very little or no impact on predicted scores; responses employing shorter and simpler words and sentences generally received comparable scores to those using more complex and diverse word and sentence structure. In contrast, previous studies which made use of human raters found large differences in scoring for texts with and without spelling errors~\cite{choi2018impact}. AI-based scoring systems may, then, offer advantages over human rating in ignoring spelling errors and writing sophistication, a particularly important conclusion for SJTs and other assessments of complex constructs where writing sophistication is typically construct-irrelevant.

Lastly, we found that off-topic responses (i.e., responses that did not relate to the content of the question asked) generally received large penalties from the scoring system and were frequently assigned the lowest possible score. Though not the main purpose of this study, we also found support for the claim that the scoring system investigated here considers different competency-related features for different items: the kinds of responses that would receive a high score for one item will not necessarily receive a high score for a different item where different features are evaluated.

\subsection{Limitations and Future Work}

This study only investigated one LLM-based scoring system for one assessment. We suggest extending this work to other LLM-based scoring systems, particularly those that use LLMs in different capacities. In particular, the system investigated here used a dual-architecture system employing LLMs for feature extraction and maintaining traditional regression techniques for feature weighting. We found, for instance, that this scoring system was robust to the evaluation of responses with spelling errors, a result that aligned with previous research~\cite{gan2024reasoning,horbach2017influence}. Those same studies provide evidence, however, that our conclusions may not extend to LLM-based systems employing less capable models (we used GPT-5.X models here)~\cite{gan2024reasoning}. Extending this research to other systems can further our understanding of the advantages and limitations of various LLM-based scoring techniques. We also suggest all developers of automated scoring systems (not just those that are LLM-based) adopt a similar approach to the one demonstrated here to investigate and categorize the influence of construct-irrelevant factors on automated scoring systems to build public trust in those systems.

In this study, we did note one particular construct-irrelevant factor that could influence the behavior of our automated scoring system: using memorized, formulaic phrases. We only investigated one such phrase here and found only a very small influence on predicted scores. We noted previously that these kinds of phrases on their own may contain some construct-relevant information, but when used frequently and superficially may be indicative of a test taker acting in bad-faith. Future work should more deeply investigate the use of formulaic phrases for our assessment and scoring system to understand the extent that these phrases can influence the predictions of the scoring system.

Even when automated scoring systems can be demonstrated to be robust to certain construct-irrelevant factors, there is still value in developing systems to flag ``unusual'' responses~\cite{higgins2006identifying,zhang2016evaluating}. While our scoring system frequently assigned the lowest available score to off-topic responses, for instance, it did not assign all such responses the lowest available score. So while it would generally not be advantageous to a test taker to provide an off-topic response, we might wish to specifically identify such responses either to ensure they are assigned a specific score or to provide additional context for decision-makers.

\begin{credits}
\subsubsection{\ackname} We would like to acknowledge Colleen Robb, Gill Sitarenios, and Josh Moskowitz for feedback on this paper.

\subsubsection{\discintname}
Both authors are employees of Acuity Insights, which administers the assessment used for data collection in this study.
\end{credits}

\bibliographystyle{splncs04}
\bibliography{bibliography}

@article{iqbal_formative,
    author = "M. Zafar Iqbal AND Rodica Ivan AND Kaitlin Bynkoski AND Kristen Archbell AND Cynthia Richard AND Kristina H. Petersen",
    title = "Formative SJT: Developing student personal and professional competencies",
    journal = "The Score",
    year = 2025
}

@inproceedings{walsh-etal-2025-using,
    title = "Using {LLM}s to identify features of personal and professional skills in an open-response situational judgment test",
    author = "Walsh, Cole  and Ivan, Rodica  and Iqbal, Muhammad Zafar  and Robb, Colleen",
    booktitle = "Proceedings of the Artificial Intelligence in Measurement and Education Conference (AIME-Con): Full Papers",
    year = "2025"
}

@inproceedings{bruno-becker-2025-explainable,
    title = "Explainable Writing Scores via Fine-grained, {LLM}-Generated Features",
    author = "Bruno, James V  and Becker, Lee",
    booktitle = "Proceedings of the Artificial Intelligence in Measurement and Education Conference (AIME-Con): Works in Progress",
    year = "2025"
}

@inproceedings{ravindran-choi-2025-investigating,
    title = "Investigating Adversarial Robustness in {LLM}-based {AES}",
    author = "Ravindran, Renjith  and Choi, Ikkyu",
    booktitle = "Proceedings of the Artificial Intelligence in Measurement and Education Conference (AIME-Con): Coordinated Session Papers",
    year = "2025"
}

@article{thomas1975test,
  title={Test-retest and inter-analyst reliability of the automated readability index, Flesch reading ease score, and the fog count},
  author={Thomas, Georgelle and Hartley, R Derald and Kincaid, J Peter},
  journal={Journal of Reading Behavior},
  volume={7},
  number={2},
  pages={149--154},
  year={1975},
  publisher={SAGE Publications Sage CA: Los Angeles, CA}
}

@inproceedings{klebanov2020automated,
  title={Automated evaluation of writing--50 years and counting},
  author={Klebanov, Beata Beigman and Madnani, Nitin},
  booktitle={Proceedings of the 58th annual meeting of the association for computational linguistics},
  pages={7796--7810},
  year={2020}
}

@article{powers2002stumping,
  title={Stumping e-rater: challenging the validity of automated essay scoring},
  author={Powers, Donald E and Burstein, Jill C and Chodorow, Martin and Fowles, Mary E and Kukich, Karen},
  journal={Computers in Human Behavior},
  volume={18},
  number={2},
  pages={103--134},
  year={2002},
  publisher={Elsevier}
}

@article{zhang2016evaluating,
  title={Evaluating the Advisory Flags and Machine Scoring Difficulty in the e-rater{\textregistered} Automated Scoring Engine},
  author={Zhang, Mo and Chen, Jing and Ruan, Chunyi},
  journal={ETS Research Report Series},
  volume={2016},
  number={2},
  pages={1--14},
  year={2016},
  publisher={Wiley Online Library}
}

@article{higgins2006identifying,
  title={Identifying off-topic student essays without topic-specific training data},
  author={Higgins, Derrick and Burstein, Jill and Attali, Yigal},
  journal={Natural Language Engineering},
  volume={12},
  number={2},
  pages={145--159},
  year={2006},
  publisher={Cambridge University Press}
}

@article{lee2024applying,
  title={Applying large language models and chain-of-thought for automatic scoring},
  author={Lee, Gyeong-Geon and Latif, Ehsan and Wu, Xuansheng and Liu, Ninghao and Zhai, Xiaoming},
  journal={Computers and Education: Artificial Intelligence},
  volume={6},
  pages={100213},
  year={2024},
  publisher={Elsevier}
}

@inproceedings{huang-wilson-2025-evaluating,
    title = "Evaluating {LLM}-Based Automated Essay Scoring: Accuracy, Fairness, and Validity",
    author = "Huang, Yue  and Wilson, Joshua",
    booktitle = "Proceedings of the Artificial Intelligence in Measurement and Education Conference (AIME-Con): Works in Progress",
    year = "2025"
}

@article{farquhar2024detecting,
  title={Detecting hallucinations in large language models using semantic entropy},
  author={Farquhar, Sebastian and Kossen, Jannik and Kuhn, Lorenz and Gal, Yarin},
  journal={Nature},
  volume={630},
  number={8017},
  pages={625--630},
  year={2024},
  publisher={Nature Publishing Group UK London}
}

@article{mcdaniel2007situational,
  title={Situational judgment tests, response instructions, and validity: A meta-analysis},
  author={McDaniel, Michael A and Hartman, Nathan S and Whetzel, Deborah L and GRUBB III, W LEE},
  journal={Personnel psychology},
  volume={60},
  number={1},
  pages={63--91},
  year={2007},
  publisher={Wiley Online Library}
}

@inproceedings{horbach2017influence,
  title={The influence of spelling errors on content scoring performance},
  author={Horbach, Andrea and Ding, Yuning and Zesch, Torsten},
  booktitle={Proceedings of the 4th workshop on natural language processing techniques for educational applications (nlptea 2017)},
  pages={45--53},
  year={2017}
}

@article{choi2018impact,
  title={The impact of spelling errors on trained raters’ scoring decisions},
  author={Choi, Ikkyu and Cho, Yeonsuk},
  journal={Language Education \& Assessment},
  volume={1},
  number={2},
  pages={45--58},
  year={2018}
}

@article{iqbal2025evaluating,
  title={Evaluating factors that impact scoring an open response situational judgment test: a mixed methods approach},
  author={Iqbal, Muhammad Zafar and Ivan, Rodica and Robb, Colleen and Derby, Jillian},
  journal={Frontiers in Medicine},
  volume={11},
  pages={1525156},
  year={2025},
  publisher={Frontiers Media SA}
}

@techreport{naismith2025duolingo,
    author = {Naismith, Ben and Cardwell, Ramsey and LaFlair, Geoffrey T. and Nydick, Steven and Kostromitina, Masha},
    title = {Duolingo english test: technical manual},
    institution = {Duolingo, Inc.},
    year = {2025}
}

@inproceedings{gan2024reasoning,
  title={Reasoning robustness of llms to adversarial typographical errors},
  author={Gan, Esther and Zhao, Yiran and Cheng, Liying and Yancan, Mao and Goyal, Anirudh and Kawaguchi, Kenji and Kan, Min-Yen and Shieh, Michael},
  booktitle={Proceedings of the 2024 Conference on Empirical Methods in Natural Language Processing},
  pages={10449--10459},
  year={2024}
}
\end{document}